\documentclass{article}
\usepackage{spconf,amsmath,graphicx}
\usepackage{algorithmic}
\usepackage{algorithm}
\usepackage{amsmath}
\usepackage{amssymb}
\usepackage{breqn}
\usepackage{bm}
\usepackage{siunitx}

\makeatletter
\def\blfootnote{\xdef\@thefnmark{}\@footnotetext}
\makeatother

\title{Reconstructing neuronal anatomy from whole-brain images \vspace{-1ex}}
\name{James Gornet$^{1, \star}$, Kannan Umadevi
  Venkataraju$^{2}$, Arun Narasimhan$^{2}$, Nicholas Turner$^{3}$}
  {Kisuk Lee$^{4}$,
  H. Sebastian Seung$^{3}$, Pavel Osten$^{2}$, Uygar
  S\"{u}mb\"{u}l$^{5}$
  \vspace{-1ex}
  }

\address{$^{1}$ \normalsize Columbia University, Department of Biomedical
  Engineering, New York, NY, USA \\
  $^{2}$ \normalsize Cold Spring Harbor Laboratory, Department of Neuroscience,
  Cold Spring Harbor, NY, USA \\
  $^{3}$ \normalsize Princeton Neuroscience Institute and Computer Science Department,
  Princeton, NJ, USA \\
  $^{4}$ \normalsize Massachusetts Institute of Technology, Department of Brain and Cognitive Sciences, Cambridge, MA, USA \\
  $^{5}$ \normalsize Allen Institute for Brain Science, Seattle, WA, USA
  \vspace{-3ex}
  }

\begin{document}

\maketitle

\blfootnote{$^{\star}$ Corresponding author: James Gornet,
  james.gornet@columbia.edu.
  This research is supported through a grant from the National Institutes
  of Health (NIMH U01MH114824). Work performed at the Allen Institute.}

\begin{abstract}
  Reconstructing multiple molecularly defined neurons from individual
  brains and across multiple brain regions can reveal organizational
  principles of the nervous system. However, high resolution imaging
  of the whole brain is a technically challenging and slow
  process. Recently, oblique light sheet microscopy has emerged as a
  rapid imaging method that can provide whole brain fluorescence
  microscopy at a voxel size of 0.4 $\times$ 0.4
  $\times$ \SI{2.5}{\micro\meter^3}. 
  On the other hand, complex image artifacts due to whole-brain
  coverage produce apparent discontinuities in neuronal arbors. Here,
  we present connectivity-preserving methods and data augmentation
  strategies for supervised learning of neuroanatomy from light microscopy using neural
  networks. We quantify the merit of our approach by implementing an
  end-to-end automated tracing pipeline. 
  Lastly, we demonstrate a
  scalable, distributed implementation that can reconstruct the large
  datasets that sub-micron whole-brain images produce.
\end{abstract}

\begin{keywords}
  image segmentation, light microscopy, machine learning
\end{keywords}

\section{Introduction}
\label{sec:intro}

Understanding the principles guiding neuronal organization has been a major goal in
neuroscience. The ability to reconstruct individual neuronal arbors is necessary, but not sufficient to achieve this goal: understanding how neurons of the same and different types co-locate themselves requires the reconstruction of the arbors of multiple neurons sharing similar molecular and/or physiological features from the same brain. Such denser reconstructions may allow the field to answer some of the fundamental questions of neuroanatomy: do cells of the same type tile across the lateral dimensions by avoiding each other? To what extent do the organizational principles within a brain region extend across the whole brain? While dense reconstruction of electron microscopy images provides a solution~\cite{denk2004serial, helmstaedter2013connectomic}, its field-of-view has been limited for studying region-wide and brain-wide organization.

Recent advances in tissue clearing~\cite{chung2013structural, susaki2014whole} and light microscopy enable a fast, and versatile approach to this problem. In particular, oblique light-sheet microscopy can image thousands of individual neurons at once from the entire mouse brain at a 0.406 $\times$ 0.406 $\times$ \SI{2.5}{\micro\meter^3} resolution~\cite{narasimhan17}. Moreover, by registering reconstructed neurons from multiple brains of different neuronal gene expressions to a common coordinate framework such as the Allen Mouse Brain Atlas~\cite{lein2007genome}, it is possible to study neuronal structure and organization across many brain regions and neuronal cell classes. Therefore, this method may soon produce hundreds of full brain images, each containing hundreds of sparsely labeled neurons. However, scaling neuronal reconstructions to such large sets is not trivial. The gold standard of manual reconstruction is a tedious and labor-intensive process with a single neuronal
reconstruction taking a few hours. This makes automated reconstruction the most viable alternative. Recently, many automated methods appeared for the reconstruction of neurons from light microscopy images. These include methods based on supervised learning with neuronal networks as well as other approaches~\cite{peng2017automatic, turetken2011automated, wang2011broadly, turetken2012automated, uygar14, gala2014active}. Some common problems include slow training and/or reconstruction speeds, tendency for topological mistakes despite high voxel-wise accuracy, and vulnerability to rare but important imaging artifacts such as stitching misalignments and microscope stage jumps. Here, we propose a supervised learning method based on a convolutional neural network architecture to address these shortcomings. In particular, we suggest (i) an objective function that penalizes topological errors more heavily, (ii) a data augmentation framework to increase robustness against multiple imaging artifacts, and (iii) a distributed scheme for scalability. Training data
augmentation for addressing microscopy image defects was initially demonstrated for automated tracing of neurons in electron microscopy images~\cite{lee2017superhuman}. Here, we adapt this approach to sparse light microscopy images.

The U-Net architecture~\cite{ronneberger15, cciccek20163d} has recently received significant interest, especially in the analysis of biomedical images. By segmenting all the voxels of an input patch rather than a central portion of it, the U-Net can learn robust segmentation rules faster, and decreases the memory and storage requirements. In this paper, we train a 3D U-Net convolutional network on a set of manually traced neuronal arbors. To overcome challenges caused by artifacts producing apparent discontinuities in the arbors, we propose a fast, connectivity-based regularization technique. While approaches that increase topological consistency exist~\cite{briggman2009maximin, jain2010boundary}, they are either too slow for peta-scale images, or are not part of an online training procedure. Our approach is a simple, differentiable modification of the cost function, and the computational overhead scales linearly with the voxel count of the input patch. On the other hand, while these regularization techniques can enforce proper connectivity, there are relatively few examples of the various imaging artifacts in the training set. In order to increase the examples of such artifacts, we simulate them through various data augmentations and present these simulations under a unified framework. Taken together, our approach produces a significant increase in the topological accuracy of neuronal reconstructions on a test set.

In addition to accuracy, an efficient, scalable implementation is necessary for reconstructing petavoxel-sized image datasets. We maintain scalability and increase the throughput by using a distributed framework for reconstructing neurons from brain images, in which the computation can be distributed across multiple GPU instances. Finally, we augment data at run-time to avoid
memory issues and computational bottlenecks. This significantly increases the throughput rate because data transfers are a substantial bottleneck. We report segmentation speeds exceeding 300 gigavoxels per hour and linear speedups in the presence of additional GPUs.

\section{Methods}
\label{sec:methods}

\subsection{Convolutional neural network regularization through
  digital topology techniques}
\label{sec:regularization}
To create the training set, we obtain volumetric reconstructions of the manual arbor traces of neuronal images by a topology-preserving inflation of the traces~\cite{sumbul2014automated}. We use a 3D U-Net convolutional neural network architecture~\cite{ronneberger15, cciccek20163d, lee2017superhuman} to learn to segment the neurons from this volumetric training set. Since neuronal morphology is ultimately represented and analyzed as a tree structure, we consider the branching pattern of the segmented neuron more important than its voxelwise accuracy. Hence, to penalize topological changes between the ground-truth and the prediction at the time of training, we binarize the network output by thresholding and identify all non-simple points in this binarized patch based on $26$-connectivity~\cite{bertrand1994new} --- points when added or removed change an object's topology (e.g., splits and mergers) --- and assign larger weights to them in the binary cross-entropy cost function
\vspace{-.1cm}
\begin{equation}
  \label{equ:loss}
  J(\hat{y}, y) = -\frac{1}{N}\sum^{N}_{i=1} w_{i} \big[ y_{i} \log
                  (\hat{y}_{i}) + (1 - y_{i}) \log (1 - {\hat{y}_{i}})
                  \big]
\end{equation}
where $w_i=w>1$ if voxel $i$ is non-simple while $w_i=1$ otherwise, $N$ is the number of voxels, and $y_i$ and $\hat{y_i}$ are the label image and predicted segmentation, respectively.
Note that the simple-ness of a voxel depends only on its $26$-neighborhood, and therefore this operation scales linearly with the patch size.

\subsection{Simulation of image artifacts through data augmentations}
\label{sec:augmentations}

\begin{figure*}[htb]
  \centering
  \includegraphics[width=178.0mm]{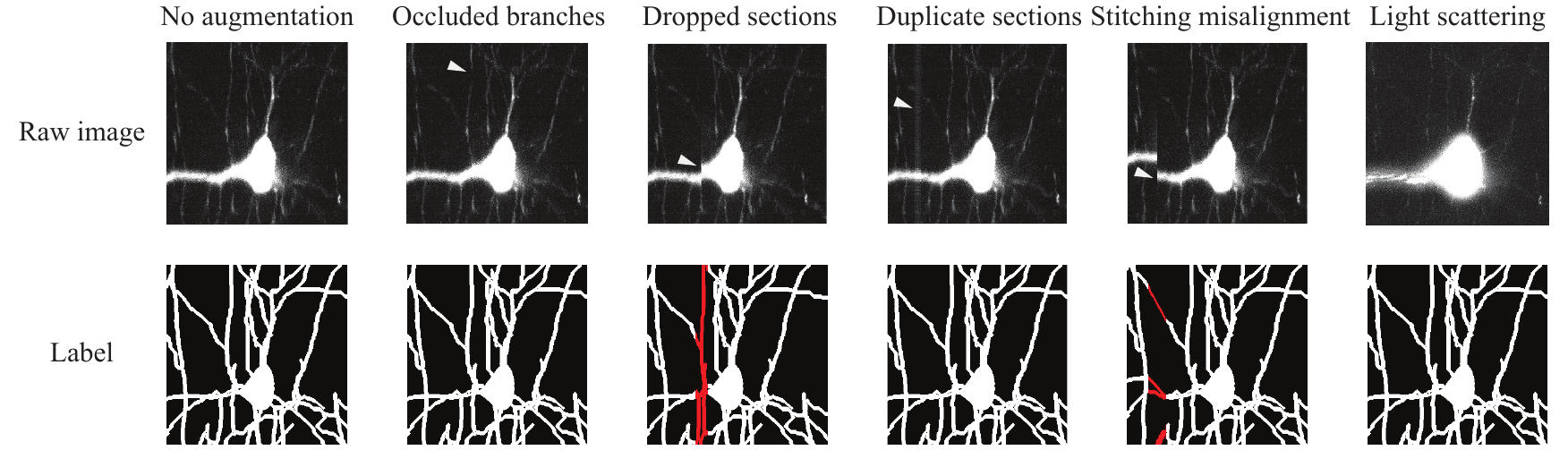}
  \caption{Data augmentations. (From left to right) No augmentation
    provides no augmentation on the raw image or the
    ground-truth. Occluded branches simulates a loss of a localized
    signal due to lack of fluorescence. Stitching misalignment
    simulates a stitching misalignment between two image
    volumes. Light scattering simulates a blurred image due to light
    scattering in the cleared tissue. Duplicate sections simulates a
    halt of the stage and an imaging of duplicate sections. Dropped
    sections simulates a jump of the stage and a missing image
    section. Artifacts in the raw images are identified by arrows while the corresponding changes in the labels are identified in red. \vspace{-2ex}}
  \label{fig:augmentations}
\end{figure*}

Data augmentation is a technique that augments the base training data with pre-defined transformations of it. By creating statistical invariances (e.g. against rotation) within the dataset or over-representing rarely occurring artifacts, augmentation can increase the robustness of the learned algorithm. Motivated by the fact that 3D microscopy is prone to several image artifacts, we followed a unified framework for data augmentation. In particular, our formalism requires explicit models of the underlying artifacts and the desired reconstruction in their presence to augment the original training set with simulations of these artifacts. 

We define the class of ``artifact-generating'' transformations as $S$ such that if $\mathcal{T} \in S$, then $\mathcal{T} = \mathcal{T}_{R} \otimes \mathcal{T}_{L}$ for
$\mathcal{T}_{R}: \mathbb{R}^{n_1 \times n_2 \times n_3} \rightarrow \mathbb{R}^{n_1 \times n_2 \times n_3}$ and
$\mathcal{T}_{L}: {\{0, 1\}}^{n_1 \times n_2 \times n_3} \rightarrow
  {\{0, 1\}}^{n_1 \times n_2 \times n_3}$, where $\mathcal{T}_R$ acts on an $n_1 \times n_2 \times n_3$ raw image and $\mathcal{T}_L$ acts on its corresponding label image. For example, the common augmentation step of rotation by $90^\circ$ can be realized by $\mathcal{T}_R$ and $\mathcal{T}_L$ both rotating their arguments by $90^\circ$. Data augmentation adds these rotated raw/label image pairs to the original training set (Fig.~\ref{fig:augmentations}).
  

{\bf Occluded branches:} Branch occlusions can be caused by
photobleaching or an absence of a fluorophore. We model the artifact-generating transformation for an absence of a fluorophore as $\mathcal{T} = \mathcal{T}_{R} \otimes \mathcal{I}$, where
\begin{equation}
    \mathcal{T_{R}}(R; x, y, z) = R - \mathrm{PSF}(x, y, z)
\end{equation}
such that $\mathcal{I}$ denotes the identity transformation, $x$ denotes the position of
the absent fluorophore and
$\mathrm{PSF}$ is its corresponding point-spread function. Here, we approximated the $\mathrm{PSF}$ of a fluorophore with a multivariate Gaussian.

{\bf Duplicate sections:} The stage of a scanning 3D microscope can
intermittently stall, which can duplicate the imaging of a tissue section. The artifact-generating transformation for stage stalling is given by $\mathcal{T} =
\mathcal{T}_{R} \otimes \mathcal{I}$, where
\begin{equation}
  \mathcal{T_R}(R; \mathbf{r}, \mathbf{r_0}, N)
  =
    \begin{cases}
      R(\mathbf{r}), & \mathbf{r} \not\in N \\
      R(\mathbf{r_{0}}), & \mathbf{r} \in N
    \end{cases}
\end{equation}
for the region $\mathbf{r} = (x, y, z)$ and the plane $\mathbf{r_0} = (x_0, y, z)$
such that $\mathcal{T}_{R}$ duplicates the slice $\mathbf{r_{0}}$ in a rectangular neighborhood $N$.

{\bf Dropped sections:} Similar to the stalling of the stage, jumps that result in missed sections can occur intermittently. The corresponding artifact-generating transformation is given by $\mathcal{T} = \mathcal{T}_{R} \otimes
\mathcal{T}_{L}$, where
\begin{equation}
  \mathcal{T}_{R}(R; \mathbf{r}, x_0, \Delta)
  =
    \begin{cases}
      R(x, y, z), & x \leq x_0 \\
      R(x + \Delta, y, z), & x > x_0
    \end{cases}
\end{equation}
and
\begin{equation}
  \mathcal{T}_{L}(L; \mathbf{r}, x_0, \Delta)
  =
    \begin{cases}
      L(x, y, z), & x \leq x_0 - \Delta \\
      L(D(x), y, z), & |x-x_0-\frac{\Delta}{2}| > \frac{3\Delta}{2} \\
      L(x + 2\Delta, y, z), & x \geq x_0 + 2\Delta
    \end{cases}
\end{equation}
such that $\mathbf{r}=(x, y, z)$, for $D(x, x_0, \Delta) = x_0 - \Delta + \frac{3}{2}\lceil x -
x_0 + \Delta \rceil$, which downsamples
the region to maintain partial connectivity in the label. Hence, $\mathcal{T}_{R}$ skips a small region given by $\Delta$ at $x_0$, and $\mathcal{T}_{L}$ is the corresponding desired transformation on the label image.

{\bf Stitching misalignment:} Misalignments can occur between 3D image
stacks, potentially causing topological breaks and mergers between
neuronal branches. The corresponding artifact-generating transformation is given by $\mathcal{T} = \mathcal{T}_{R} \otimes
\mathcal{T}_{L}$, where
\begin{equation}
  \mathcal{T}_{R}(R;x,y,z,\Delta)
  =
    \begin{cases}
      R(x, y, z), & x \leq x_{0} \\
      R(x, y + \Delta, z), & x > x_{0}
    \end{cases}
\end{equation}
and
\begin{equation}
  \mathcal{T}_{L}(L;x,y,z,\Delta)
  = 
    \begin{cases}
      L, & x \leq x_{0} - \frac{1}{2} \Delta \\
      \Sigma_{zy}(\Delta) L, & |x-x_{0}|<\frac{1}{2} \Delta \\
      L(x, y + \Delta, z), & x > x_{0} + \frac{1}{2} \Delta
    \end{cases}
\end{equation}
such that $\Sigma_{zy}(\Delta)$ is a shear transform on $L$.  Hence,
$\mathcal{T}_{R}$ translates a region of $R$ to simulate a stitching
misalignment, and $\mathcal{T}_{L}$ shears a region around the
discontinuity to maintain 18-connectivity in the label.

{\bf Light scattering:} Light scattering by the cleared tissue can create an
inhomogeneous intensity profile and blur the image. To simulate this
image artifact, we assumed the scatter has a homogeneous profile and is anisotropic due to the oblique light-sheet. We approximate these characteristics with a Gaussian kernel:
$G(x,y,z)=G(\mathbf{r})=\mathcal{N}(\mathbf{r}; \mu, \Sigma)$. In addition, the global inhomogeneous intensity profile was simulated with an additive constant. Thus, the corresponding artifact-generating transformation is given by $\mathcal{T} = \mathcal{T}_{R} \otimes
\mathcal{I}$, where
\begin{equation}
\mathcal{T}_{R}(R) = R(x,y,z)*G(x,y,z)+\lambda
\end{equation}

\subsection{Fully automated, scalable tracing}
\label{sec:system}

To optimize the pipeline for scalability, we store images as parcellated HDF5 datasets.
For training, a file server software streams these images to the GPU server, which performs data augmentations on-the-fly, to minimize storage space requirements. For deploying the trained neural network, the file server similarly streams the datasets to a GPU server for segmentation. Once the segmentation is completed, the neuronal morphology is reconstructed automatically from the segmented image using the UltraTracer neuron tracing tool within the Vaa3D software package~\cite{peng2017automatic}.

\section{Experimental Procedure}
\label{sec:experiment}
In our experiments, we used a dataset of 54 manually traced neurons
imaged using oblique light-sheet microscopy. These morphological
annotations were dilated while preserving topology for training the
neural network for segmentation. We partitioned the dataset into
training, validation, and test sets by randomly choosing 25, 8, and 21 neurons, respectively. The
software package PyTorch was used to implement the neural
network~\cite{paszke17}.  The network was trained using an Adam
optimizer for gradient descent~\cite{kingma15}. Training and
reconstruction were conducted on two Intel Xeon Silver 4116 CPU, 256 GB
RAM, and 2 NVIDIA GeForce GTX 1080 Ti GPUs.

\vspace{-2ex}
\section{Results}
\label{sec:results}

\subsection{Topologically accurate reconstruction}
\label{sec:cortex}

\begin{figure}[htb]
  \centering
  \includegraphics[width=86.0mm]{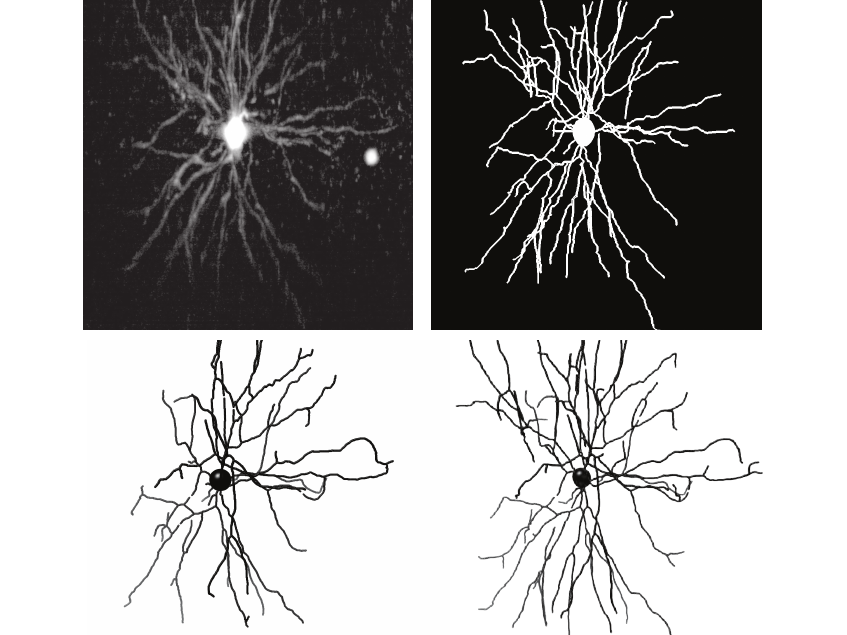}
  \caption{Neuronal
    images reconstructed using the U-Net architecture. (Upper-left) Raw image. (Upper-right) Label. (Lower-left) Reconstruction performed without augmentations or regularization. (Lower-right)  Reconstruction performed with augmentations and regularization.
    \vspace{-2ex}}
\end{figure}

\begin{figure}[htb]
  \centering
  \includegraphics[width=80.0mm]{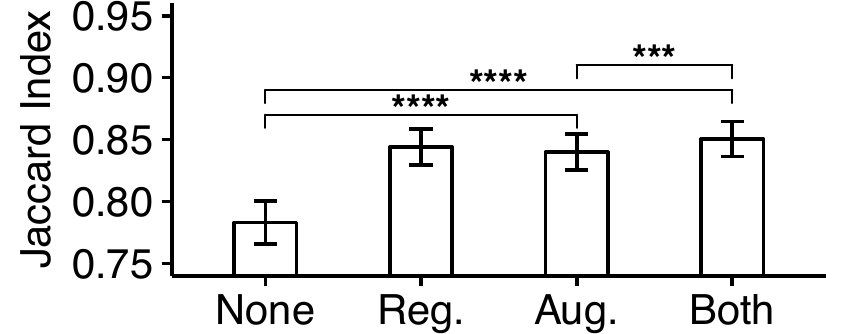}
  \caption{Evaluation of segmentation results. The groups None,
    Reg., Aug., and Both represent the trials with
    no augmentations or regularization, with the connectivity-based
    regularization, with augmentations, and with both augmentations and regularization,
    respectively. The groups were quantified using the Jaccard index
    and compared using a paired Student's t-test. ({\small$\star\star\star$} indicates $p < 0.001$ and {\small$\star\star\star\star$} indicates $p < 0.0001$) 
    \vspace{-2ex}}
  \label{fig:results}
\end{figure}

	To quantify the topological accuracy of the network on light-sheet microscopy data, we define the topological error as the number of non-simple points that must be added or removed from a prediction to obtain its corresponding label. Specifically, for binary images $\hat{L}$ and $L$, let $\mathcal{W}(\hat{L},L)$ denote a topology-preserving warping of $\hat{L}$ that minimizes the voxelwise disagreements between the warped image and $L$~\cite{jain2010boundary, uygar14}, $\hat{L}\cap L$ denote the binary image whose foreground is common to both $\hat{L}$ and $L$, and $c(L)$ denote the number of foreground voxels of $L$. We quantify the agreement between a reconstruction $\hat{L}$ and label $L$ using the Jaccard index as
	\begin{equation}
J(\hat{L}, L) = \frac{c(\mathcal{W}(\hat{L}, L) \cap L)}{c(\mathcal{W}(\hat{L}, L) \cup L)}.
    \end{equation}
We compared this score across different U-Net results: without any augmentations or regularization, with the augmentations, with the topological regularization, and with both the topological regularization and the augmentations. The U-Net results with augmentations and topological regularization performed significantly better compared to the results without augmentations or regularization (Figs 2, 3).


\subsection{Neuron reconstruction is efficient and scalable}
\label{sec:efficient}

To quantify the efficiency of the distributed framework, we measured
the framework’s throughput for augmenting data, training on the data,
and segmenting the data. Augmentations performed at 35.2 $\pm$ 9.2 gigavoxels per hour while training
performed at 16.8 $\pm$ 0.2 megavoxels per hour. Segmentation performed at 348.8 $\pm$ 1.9 gigavoxels per
hour. Both segmentation and training showed a linear speedup
with an additional GPU. For an entire mouse brain, neuronal reconstruction would take about 23 hours on a single GPU.


\section{Discussion}
\label{sec:discussion}

In this paper, we proposed an efficient, scalable, and accurate algorithm capable of reconstructing neuronal anatomy from light microscopy images of the whole brain. Our method employs topological regularization as well as simulates discontinuous image artifacts inherent to the imaging systems. These techniques help maintain topological correctness of the trace (skeleton) representations of neuronal arbors. 

While we demonstrated the merit of our approach on neuronal images obtained by oblique light-sheet microscopy, our methods address some of the problems common to most 3D fluorescence microscopy techniques. Therefore, we hope that some of our methods will be useful for multiple applications. Combined with the speed and precision of oblique light-sheet microscopy, the distributed and fast nature of our approach enables the production of a comprehensive database of neuronal anatomy across many brain regions and cell classes. We believe that these aspects will be useful in discovering different cortical cell types as well as understanding the anatomical organization of the brain.

\newpage
\begin{small}
\bibliographystyle{IEEEbib}
\bibliography{paper}

\begin{thebibliography}{10}

\bibitem{denk2004serial}
Winfried Denk and Heinz Horstmann,
\newblock ``Serial block-face scanning electron microscopy to reconstruct
  three-dimensional tissue nanostructure,''
\newblock {\em PLoS Biology}, vol. 2, no. 11, pp. e329, 2004.

\bibitem{helmstaedter2013connectomic}
Moritz Helmstaedter, Kevin~L. Briggman, Srinivas~C. Turaga, Viren Jain,
  H.~Sebastian Seung, and Winfried Denk,
\newblock ``Connectomic reconstruction of the inner plexiform layer in the
  mouse retina,''
\newblock {\em Nature}, vol. 500, no. 7461, pp. 168, 2013.

\bibitem{chung2013structural}
Kwanghun Chung, Jenelle Wallace, Sung-Yon Kim, Sandhiya Kalyanasundaram,
  Aaron~S. Andalman, Thomas~J. Davidson, Julie~J. Mirzabekov, Kelly~A.
  Zalocusky, Joanna Mattis, Aleksandra~K. Denisin, et~al.,
\newblock ``Structural and molecular interrogation of intact biological
  systems,''
\newblock {\em Nature}, vol. 497, no. 7449, pp. 332, 2013.

\bibitem{susaki2014whole}
Etsuo~A. Susaki, Kazuki Tainaka, Dimitri Perrin, Fumiaki Kishino, Takehiro
  Tawara, Tomonobu~M. Watanabe, Chihiro Yokoyama, Hirotaka Onoe, Megumi Eguchi,
  Shun Yamaguchi, et~al.,
\newblock ``Whole-brain imaging with single-cell resolution using chemical
  cocktails and computational analysis,''
\newblock {\em Cell}, vol. 157, no. 3, pp. 726--739, 2014.

\bibitem{narasimhan17}
Arun Narasimhan, Kannan~Umadevi Venkataraju, Judith Mizrachi, Dinu~F. Albeanu,
  and Pavel Osten,
\newblock ``Oblique light-sheet tomography: fast and high resolution volumetric
  imaging of mouse brains,''
\newblock {\em BioRxiv}, 2017.

\bibitem{lein2007genome}
Ed~S. Lein, Michael~J. Hawrylycz, Nancy Ao, Mikael Ayres, Amy Bensinger, Amy
  Bernard, Andrew~F. Boe, Mark~S. Boguski, Kevin~S. Brockway, Emi~J. Byrnes,
  et~al.,
\newblock ``Genome-wide atlas of gene expression in the adult mouse brain,''
\newblock {\em Nature}, vol. 445, no. 7124, pp. 168, 2007.

\bibitem{peng2017automatic}
Hanchuan Peng, Zhi Zhou, Erik Meijering, Ting Zhao, Giorgio~A. Ascoli, and
  Michael Hawrylycz,
\newblock ``Automatic tracing of ultra-volumes of neuronal images,''
\newblock {\em Nature Methods}, vol. 14, no. 4, pp. 332, 2017.

\bibitem{turetken2011automated}
Engin T{\"u}retken, Germ{\'a}n Gonz{\'a}lez, Christian Blum, and Pascal Fua,
\newblock ``Automated reconstruction of dendritic and axonal trees by global
  optimization with geometric priors,''
\newblock {\em Neuroinformatics}, vol. 9, no. 2-3, pp. 279--302, 2011.

\bibitem{wang2011broadly}
Yu~Wang, Arunachalam Narayanaswamy, Chia-Ling Tsai, and Badrinath Roysam,
\newblock ``A broadly applicable {3-D} neuron tracing method based on
  open-curve snake,''
\newblock {\em Neuroinformatics}, vol. 9, no. 2-3, pp. 193--217, 2011.

\bibitem{turetken2012automated}
Engin T{\"u}retken, Fethallah Benmansour, and Pascal Fua,
\newblock ``Automated reconstruction of tree structures using path classifiers
  and mixed integer programming,''
\newblock in {\em Computer Vision and Pattern Recognition (CVPR), 2012 IEEE
  Conference on}. IEEE, 2012, pp. 566--573.

\bibitem{uygar14}
Uygar S\"{u}mb\"{u}l, Sen Song, Kyle McCulloch, Michael Becker, Bin Lin,
  Joshua~R. Sanes, Richard~H. Masland, and H.~Sebastian Seung,
\newblock ``A genetic and computational approach to structurally classify
  neuronal types,''
\newblock {\em Nature Communications}, vol. 5, no. 3512, 2014.

\bibitem{gala2014active}
Rohan Gala, Julio Chapeton, Jayant Jitesh, Chintan Bhavsar, and Armen
  Stepanyants,
\newblock ``Active learning of neuron morphology for accurate automated tracing
  of neurites,''
\newblock {\em Frontiers in Neuroanatomy}, vol. 8, pp. 37, 2014.

\bibitem{lee2017superhuman}
Kisuk Lee, Jonathan Zung, Peter Li, Viren Jain, and H.~Sebastian Seung,
\newblock ``Superhuman accuracy on the {SNEMI3D} connectomics challenge,''
\newblock {\em ArXiv}, May 2017,
\newblock arXiv:1706.00120 [cs].

\bibitem{ronneberger15}
Olaf Ronneberger, Philipp Fischer, and Thomas Brox,
\newblock ``{U-Net}: Convolutional networks for biomedical image
  segmentation,''
\newblock in {\em Medical Image Computing and Computer-Assisted Intervention --
  MICCAI 2015}, Nassir Navab, Joachim Hornegger, William~M. Wells, and
  Alejandro~F. Frangi, Eds., Cham, 2015, pp. 234--241, Springer International
  Publishing.

\bibitem{cciccek20163d}
{\"O}zg{\"u}n {\c{C}}i{\c{c}}ek, Ahmed Abdulkadir, Soeren~S. Lienkamp, Thomas
  Brox, and Olaf Ronneberger,
\newblock ``{3D} {U-Net}: Learning dense volumetric segmentation from sparse
  annotation,''
\newblock in {\em International Conference on Medical Image Computing and
  Computer-Assisted Intervention}. Springer, 2016, pp. 424--432.

\bibitem{briggman2009maximin}
Kevin Briggman, Winfried Denk, H.~Sebastian Seung, Moritz~N. Helmstaedter, and
  Srinivas~C. Turaga,
\newblock ``Maximin affinity learning of image segmentation,''
\newblock in {\em Advances in Neural Information Processing Systems}, 2009, pp.
  1865--1873.

\bibitem{jain2010boundary}
Viren Jain, Benjamin Bollmann, Mark Richardson, Daniel~R. Berger, Moritz~N.
  Helmstaedter, Kevin~L. Briggman, Winfried Denk, Jared~B. Bowden, John~M.
  Mendenhall, Wickliffe~C. Abraham, et~al.,
\newblock ``Boundary learning by optimization with topological constraints,''
\newblock in {\em Computer Vision and Pattern Recognition (CVPR), 2010 IEEE
  Conference on}. IEEE, 2010, pp. 2488--2495.

\bibitem{sumbul2014automated}
Uygar S{\"u}mb{\"u}l, Aleksandar Zlateski, Ashwin Vishwanathan, Richard~H.
  Masland, and H.~Sebastian Seung,
\newblock ``Automated computation of arbor densities: a step toward identifying
  neuronal cell types,''
\newblock {\em Frontiers in Neuroanatomy}, vol. 8, pp. 139, 2014.

\bibitem{bertrand1994new}
Gilles Bertrand and Gr{\'e}goire Malandain,
\newblock ``A new characterization of three-dimensional simple points,''
\newblock {\em Pattern Recognition Letters}, vol. 15, no. 2, pp. 169--175,
  1994.

\bibitem{paszke17}
Adam Paszke, Sam Gross, Soumith Chintala, Gregory Chanan, Edward Yang, Zachary
  DeVito, Zeming Lin, Alban Desmaison, Luca Antiga, and Adam Lerer,
\newblock ``Automatic differentiation in {PyTorch},''
\newblock {\em NIPS}, 2017.

\bibitem{kingma15}
Diederik~P. Kingma and Jimmy Ba,
\newblock ``{Adam: A Method for Stochastic Optimization},''
\newblock {\em ArXiv}, Dec. 2014,
\newblock arXiv:1412.6980 [cs].

\end{thebibliography}
\end{small}
\end{document}